\documentclass[runningheads]{llncs}
\usepackage{graphicx}
\usepackage{amsmath,amssymb} % define this before the line numbering.
\usepackage{color}
\usepackage{booktabs} % For formal tables
\usepackage{floatrow}
\newfloatcommand{capbtabbox}{table}[][\FBwidth]
\usepackage{multirow}
\usepackage[font=small,labelfont=bf]{caption}
\usepackage{xcolor}
\newcommand{\kaili}[1]{\textcolor{black}{#1}}
\newcommand{\jose}[1]{\textcolor{black}{#1}}
\newcommand{\tinne}[1]{\textcolor{black}{#1}}
\newcommand{\tocheck}[1]{\textcolor{black}{#1}}

\begin{document}
\pagestyle{headings}
\mainmatter

\def\ACCV20SubNumber{559}  % Insert your submission number here

%===========================================================
\title{In Defense of LSTMs for addressing Multiple Instance Learning Problems} % Replace with your title
\titlerunning{LSTM for Multiple Instance Learning}
\authorrunning{K.Wang et al.}

\author{Kaili Wang$\dagger$, Jose Oramas*, Tinne Tuytelaars$\dagger$}
\institute{$\dagger$KU Leuven, ESAT-PSI, *University of Antwerp-imec-IDLab}

\maketitle

%===========================================================
\begin{abstract}
LSTMs have a proven track record in analyzing sequential data. 
But \tinne{what about}
%are they also useful for processing 
unordered instance bags, \kaili{as found under a Multiple Instance Learning (MIL) setting}? 
% Here,
% we investigate the potential of LSTMs at capturing information beyond order.
% Multi
\tinne{While not often used for this, we show LSTMs excell under this setting too.}
%We formulate the learning of the underlying structure within a bag of instances {using LSTM} as a MIL problem. 
In addition, we show that LSTMs are capable of indirectly capturing instance-level information
using only bag-level annotations. 
Thus, they can be used to learn instance-level models in a weakly supervised manner.
Our empirical evaluation on both simplified (MNIST) and realistic (Lookbook and Histopathology) datasets shows that 
%the proposed method is 
\tinne{LSTMs are}
competitive with or even surpass state-of-the-art methods specially designed for handling 
\tinne{specific} MIL problems. 
Moreover, we show that their performance on instance-level prediction is close to that of fully-supervised methods.
\end{abstract}

%%=====================================================================================
%%  INTRODUCTION
%%=====================================================================================

\section{Introduction}

\jose{
Traditional single-instance classification methods focus on learning a mapping between a feature 
vector (extracted from a single instance) w.r.t. a specific class label.
In a complementary fashion, Multiple Instance Learning (MIL)~\cite{MILencyclopedia} algorithms 
are tasked with learning how to associate a set of elements, usually referred to as a "bag", 
with a specific label. 
In comparison, MIL methods usually require weaker supervision in the form of bag-level labels.
%
% The capability of making predictions over groups of elements while requiring only weak supervision is
% a characteristic that makes this family of methods attractive to address several real-world applications.
% Examples include drug activity prediction, image classification~\cite{Carbonneau2018MultipleIL,MIimagesWei2016}, 
% image retrieval, sound classification~\cite{MIsoundClassification}, 
% anomaly detection~\cite{MITanomalyDetection}, medical imaging~\cite{MILbreastCancer} 
% and web-mining.
}
\tinne{The MIL problem has a long history, and various solutions have been proposed over time.
Here, we advocate the use of standard LSTM networks in this context, as a strong baseline,
yielding competitive results under a wide range of MIL settings.}

Long short-term memory (LSTM) networks~\cite{LSTM_97} have been proposed {as an extension 
over standard recurrent neural networks, to store information
over long time intervals in sequential data. 
% and avoid issues with vanishing or exploding gradients.
They have been used extensively and very successfully}
for modeling sentences (sequences of words) in text documents~\cite{Sundermeyer_lstmneural}, e.g.~for machine translation~\cite{SutskeverNIPS2014} or sentiment analysis~\cite{LeMikolovICML2014}. Later, they have been employed in several other fields including computer vision \cite{LSTM_action_recog_16,LSTM_CV_15,Alahi_2016_CVPR,zhu_aaai16,Singh_2016_CVPR} and speech processing~\cite{LSTM_speech_05,Sundermeyer_lstmneural}. 
%where they have proven effective for modeling sequences of other modalities of data as well.
%
LSTMs provide great flexibility for handling data sequences. 
%On the one hand, t
There is no need to know the length of the sequence beforehand, 
%On the other hand, 
and they can operate on sequences of variable size.
In addition, they are capable of accumulating information by using a memory mechanism with add/forget\cite{Gers99learningtoforget} functionality.

{\em As they need the input to be provided as a sequence, LSTMs do not seem an appropriate choice for analyzing unordered bags at first - but is that so ? }
%
% For the more extreme case of fine-grained actions (where appearance is very similar), the "order characteristic of the representation" might become more critical (e.g. "opening a door" vs "closing a door").
% In this setting, the loss pushes the LSTMs towards giving strong relevance to the order in which the "visual features" should occur in order to discriminate between those two actions.
% On the other extreme, i.e. actions driven by scene (e.g. "playing tennis" vs "cooking"), the simple occurrence of a given "visual feature", e.g. tennis court or racket, might be sufficient, which means that setting the "order" characteristic has close to no value in this classification part, normal CNN can tackle them \cite{GeestGGLST16}.
% It is noticeable that it is the \textit{memory ability} that makes LSTM remember the temporal (order) information.
% According to the equation, the output depends on the current input as well as the previous output. 
% This memory ability can help LSTM capture more. 
%
%On the one hand, 
%we would expect 
Obviously, 
%processing sequential data requires a mechanism to remember information from earlier elements in the sequence. As such,
the capability to remember the temporal (order) information
can be attributed to the \textit{memory} ability of LSTMs. 
%However, 
%on the other hand, 
%we would expect 
This memory ability is capable of capturing other types of information beyond order as well. 
{Take the LSTMs used for action recognition as an example. For some finegrained actions (e.g. opening or closing a door), the order of the events is key and this is picked up by the LSTM.
However, for other actions the context information provides the most important cue (e.g. playing tennis or cooking). This does not depend on the temporal order, but can still be learned using LSTM.}

\vspace{-0.4mm}

{Starting from an unordered bag, we can always transform it into a sequence by imposing
a random order on the instances, making it suitable for LSTMs. The order is not relevant,
but that does not matter: the LSTM can still process the data and extract useful information from it. 
In fact, this is also how humans often deal with unordered bags: e.g, if one is asked to count the number
of rotten apples in a basket, most of us would just pick the apples one by one, in random order, inspect them and keep track of the count. The order does not matter, but nevertheless, treating them in a sequential order comes very naturally. 

\jose{The observations above clearly hint to a promising capability of LSTMs for addressing MIL problems.}
Yet, LSTMs are not often used in this way \tinne{(see our related work section for a few notable exceptions).}}
%- except for \cite{LSTM_97,fashionCompLSTM,yazici2019orderless,NTM14,graves2016hybrid,deepset17}.}
%
%In this work, 
Therefore, 
we present a systematic analysis on the performance of LSTMs when 
% operating over unordered sets.
\kaili{addressing MIL problems.
% and its potential ability in weakly supervised learning.
}
%In what follows, we provide a systematic analysis on the performance of LSTMs when operating over unordered sets. 
More specifically, we conduct a series of experiments considering different factors that may 
%have a significant effect on 
affect the performance of LSTMs. 
First, \tinne{we consider the standard MIL problem~\cite{AmoresMILsurvey2013,Carbonneau2018MultipleIL,foulds_frank_2010}, 
with bags of instances without sequential order.}
%bags with underlying structures that differ from the traditional data sequences defined by "ordered" instances. This is formulated as a MIL problem~\cite{AmoresMILsurvey2013,Carbonneau2018MultipleIL,foulds_frank_2010}, where the goal is to perform predictions over bags of instances with more general underlying structures.
Second, we study the effect of the order in which the instances of each bag are fed to the LSTM network. 
Likewise, in a third test, we investigate the influence of the cardinality (size) of the bag on performance.
Fourth, we assess the effect that the complexity of the data has on the previous observations. Toward this goal we conduct experiments considering bags derived from the MNIST dataset \cite{lecun-MNIST-2010}, clothing-item images from the Lookbook dataset \cite{Lookbook} and Histopathology images \cite{ITW:2018}.
Fifth, we inspect how the internal state of the LSTM changes when observing each of the instances of the bag.
% Moreover, we investigate to what extent changes in the internal representation might be informative for different prediction problems over sets. 
% {THIS IS UNCLEAR. IS THIS REFERRING TO THE WEAKLY SUPERVISED EXPERIMENTS ? WE SHOULD MENTION THEM HERE MORE EXPLICITLY, NOT JUST IN THE CONTRIBUTIONS BELOW.}
{Moreover, we propose an LSTM-based framework that can predict the instance-level labels by only using bag-level labels, in a weakly supervised manner.}

Our contributions are three-fold: 
% i) a novel iterative method that is capable 
% of modelling the underlying assumption / relationship characterizing the 
% elements in a bag 
% without the need of explicit heuristics,
% ii) a robust framework able to handle various typical assumptions considered 
% by MIL problems,
% and iii) a framework capable of modelling distributions at the instance-level, using only bag-level annotations.
% i) We advocate the application of LSTM networks beyond ordered sets, as a means for encoding more-general underlying structures within bags of instances.
\kaili{i) We advocate the application of LSTM networks on general MIL problems, as a means for encoding more general underlying structures (i.e. not limited to ordered data) within bags of instances.}
% ii) 
% %through an extensive evaluation, 
% We empirically show that LSTMs are capable of capturing information within bags that go beyond ordered sequences. Moreover, we show that their performance is, surprisingly, comparable or even superior to that of methods specially designed to handle sets.
\kaili{ii) 
% We empirically and systematically validate
We conduct an extensive systematic evaluation showing
that LSTMs are capable of capturing information within bags that go beyond ordered sequences. Moreover, we show that their performance is, surprisingly, comparable or even superior to that of methods especially designed to handle MIL problems.}
iii) We propose a framework for weakly supervised learning based on LSTM, capable of modeling distributions at the instance-level, using only bag-level annotations.

% {can we also say that use LSTM to solve MIL problem, like in AAAI/CVPR?}

%%==============================================================================
%%  RELATED WORK
%%==============================================================================

\section{Related Work}
\label{sec:relatedWork}
% {add the reference where LSTM is used to capture temporal data?}
%===========
%CFCM: Segmentation via Coarse to Fine Context Memory: Use LSTM to do segmentation, but the input of LSTM is still ordered.

%===========
Efforts based on LSTMs/RNNs aiming at modelling unordered bags are quite rare.
%There is rare work which utilizes LSTM or RNN models to tackle unordered data.
\kaili{\cite{NTM14,graves2016hybrid} propose a memory-based recurrent architecture and apply it on sorting \cite{NTM14} and traversal of graphs \cite{graves2016hybrid}, where the input data can be regarded as unordered bags.}
\cite{fashionCompLSTM} considers a fashion outfit to be a sequence
(from top to bottom) and each item in the outfit as an instance. Then, a LSTM model is trained to sequentially predict the next item conditioned on previous ones to learn their compatibility relationships. 
%Perse, the compatibility prediction task is not ordered, however \cite{fashionCompLSTM} introduce a notion of order to ease the application of LSTMs.
%
Later, \cite{yazici2019orderless} used a CNN-RNN model to perform multiple label predictions, where LSTMs were used to decode the labels in an unordered manner.
% by substituting the loss function by an orderless loss function. 
Different from this work which focused on the decoding part, we investigate the encoding of the unordered bags.
% \kaili{\cite{sodeep19} utilized LSTM to design a sorter with respect to the Spearman correlation metrics. However, the input of the sorter is ordered video frames. \cite{LSTM_counting_19} designed a counter using LSTM for Dyck language.}
%
\kaili{\cite{deepset17} 
% designed a permutation-invariant model to learn bag representations. It showed that 
proposed to learn permutation-invariant bag representations by summing features across instances and applying non-linear transformations. This can be regarded as a specific case of \cite{ITW:2018} where the weight of the instances are uniform. 
% Therefore, it works to some extent even without the attention mechanism from  \cite{ITW:2018}.
% \cite{deepset17} also included a LSTM-based model for a specific non-MIL experiment: predicting sum of digits as a benchmark experiment. 
% Every element in the bag is equally important and the model only needs to sum everything, therefore, even without attention mechanism \cite{ITW:2018} the model works.
}
\kaili{These works either use LSTM to handle unordered data on some specific settings~\cite{fashionCompLSTM,deepset17} or use them just as side experiments~\cite{NTM14,graves2016hybrid}. Here, we propose the use of LSTMs to address more \textit{general} MIL problems.}

%%%=======================================
% On the task of modeling general bag representations, 
% over the last decade various approaches have been proposed to 
% address several types of MIL problems. Since our work
% is based on LSTMs, we 
% position our approach w.r.t. efforts based on neural 
% networks, specifically those with deep architectures. 
% Please refer to \cite{AmoresMILsurvey2013,Carbonneau2018MultipleIL,foulds_frank_2010} 
% for detailed surveys covering non-deep methods.
%%%=======================================
\kaili{On the task of modeling general bag representations, we 
position our approach w.r.t. efforts based on neural 
networks, specifically those with deep architectures since our work
is based on LSTMs. 
Please refer to \cite{AmoresMILsurvey2013,Carbonneau2018MultipleIL,foulds_frank_2010} 
for detailed surveys covering non-deep methods.}
\cite{RamonMINN} 
% constitutes one of the first efforts towards 
% addressing MIL problems through neural networks.
% The 
proposed a multiple instance neural network to estimate 
instance probabilities 
% which are aggregated at the last layer using 
% a convex max operator in order to predict a bag probability
.
This idea is further extended in \cite{WangDeepMIL} which 
%proposes to use 
uses a neural network to learn a bag representation and directly 
carry out bag classification without estimating instance-level 
probabilities or labels. 
In parallel, \cite{ITW:2018} proposed an attention mechanism 
to learn a pooling operation over instances. 
The weights learned for the attention mechanism on instances can 
serve as indicators of the contribution of each instance to the 
final decision -- thus, producing explainable predictions.
\cite{LiuCVPR17attention} proposed a similar idea, using the
computed bag representations, to measure distances between image 
bags.
\cite{yan18DynamicPooling} proposed to update the contributions 
of the instances by observing all the instances of the 
bag a predefined number of iterations.
Along a different direction, \cite{tiboMMIN} proposed a hierarchical 
bag representation in which each bag is internally divided into 
subbags until reaching the instance level. 
Very recently, \cite{MingMIGNN} proposed to consider the instances 
in the bags to be non-i.i.d. and 
%propose the 
used graph neural networks to learn a bag embedding. 

%%%==========original==========
% Similar to \cite{ITW:2018,WangDeepMIL} we embed the instance features 
% from each bag into a common space from which a bag representation is 
% learned. This bag representation is used to make directly bag predictions 
% related to MIL problems.
% Similar to \cite{MingMIGNN} and \cite{yan18DynamicPooling} we aim 
% at learning the underlying structure within the sets. 
% %
% Different from \cite{MingMIGNN}, our method does 
% not rely on hand-tuned parameters, e.g. distance thresholds to 
% define edges in the graph, and other manual graph construction. 
% Moreover, the improvement in performance displayed by our method 
% is not sensitive to the possible lack of structure within each set.
% Compared to \cite{yan18DynamicPooling}, our method only requires 
% \textit{a single pass} through all the instances. Moreover, our 
% method is able to go beyond binary classification tasks and handle 
% more complex classification and regression tasks.
% %
% Finally, most of the works mentioned above operate under the standard 
% MI learning assumption. In contrast, the proposed  
% approach is able to learn the underlying structure of bags of instances, 
% thus, being robust to several MIL assumptions/problems~\cite{foulds_frank_2010}.
%%%%=============================
Similar to \cite{ITW:2018,WangDeepMIL} we embed the instance features 
from each bag into a common space and the bag representation is used to make direct bag predictions. 
% related to MIL problems.
Similar to \cite{MingMIGNN,yan18DynamicPooling} we aim 
at learning the underlying structure within the bags. 
Different from \cite{MingMIGNN}, our method does 
not rely on hand-tuned parameters, e.g. 
% distance thresholds to define edges in the graph, and other 
manual graph construction. 
Moreover, the improvement in performance displayed by our method 
is not sensitive to the possible lack of structure within each bag.
Compared to \cite{yan18DynamicPooling}, our method only requires 
\textit{a single pass} through all the instances. Moreover, our 
method is able to go beyond binary classification tasks and handle 
more complex classification and regression tasks.
Finally, most of the works mentioned above operate under the standard 
Multiple Instance (MI) assumption. In contrast, the proposed  
approach is able to learn the underlying structure of bags of instances, 
thus, being robust to several MI assumptions/problems~\cite{foulds_frank_2010}.

\section{Methodology}
\label{sec:methodology}
%
% We begin our analysis by defining the different instances that compose it. First, we formally define the different underlying structures that could be present on a bag of instances. Then, we introduce the LSTM-based pipeline that will be considered to model bags of instances throughout our analysis.
\kaili{We begin our analysis by defining the different parts that compose it. First, we formally define MIL problems and draw pointers towards different MI assumptions that they commonly consider.
% the different underlying structures that could be present on a bag of instances and revisit the MI Assumptions. 
Then, we introduce the LSTM-based pipeline that will be considered to model bags of instances throughout our analysis.}
%
%%--------------------------
%% Problem [MIL] formulation
%%--------------------------
%
% \subsection{Underlying Structures within bags of Instances}
\subsection{Underlying Structures within Bags of Instances 
% and MI Assumptions
}
\label{sec:MILformulation}
%
% As was said earlier, underlying sequential structures between the instances within a bag is a cue that LSTMs are capable of encoding quite effectively. In fact, this capability have made them effective at handling problems defined by these sequences, e.g. actions, speech, text. However, this sequential order is just one of many possible underlying structures that could be present between the instances within the bags processed by a LSTM.
% Encoding these underlying structures and making predictions about bags of instances is the main objective of Multiple Instance Learning (MIL)~\cite{AmoresMILsurvey2013,Carbonneau2018MultipleIL,foulds_frank_2010}. We will conduct our analysis from the perspective of MIL, where LSTMs will play an active role in modeling the underlying bag structure. 
% %
% Given the bag $X_j{=}\{x_1,x_2,...,x_m\}$ of instances $x_i$ with 
% latent instance-level labels $C_j{=}\{c_1,c_2,...,c_m\}$, traditional 
% MIL problems aim at the prediction of bag-level labels $y_j$ for each bag $X_j$. 
% The MIL literature covers several underlying bag structures, referred to as \textit{assumptions}, that have been commonly considered in order to define bag-level labels. We refer the reader to \cite{AmoresMILsurvey2013,Carbonneau2018MultipleIL,foulds_frank_2010} 
% for different surveys that have grouped these assumptions based on different criteria.
As was said earlier, underlying sequential structures between the instances within a bag is a cue that LSTMs are capable of encoding quite effectively. In fact, this capability have made them effective at handling problems defined by these sequences, e.g. actions, speech, text. However, this sequential order is just one of many possible underlying structures that could be present between the instances within the bags processed by a LSTM.
Encoding these underlying structures and making predictions about bags of instances is the main objective of Multiple Instance Learning (MIL)~\cite{AmoresMILsurvey2013,Carbonneau2018MultipleIL,foulds_frank_2010}. We will conduct our analysis from the perspective of MIL, where LSTMs will play an active role in modeling the underlying bag structure. 

Given the bag $X_j{=}\{x_1,x_2,...,x_m\}$ of instances $x_i$ with 
latent instance-level labels $C_j{=}\{c_1,c_2,...,c_m\}$, traditional 
MIL problems aim at the prediction of bag-level labels $y_j$ for each bag $X_j$. 
The MIL literature covers several underlying bag structures, referred to as \textit{assumptions}, that have been commonly considered in order to define bag-level labels. We refer the reader to \cite{AmoresMILsurvey2013,Carbonneau2018MultipleIL,foulds_frank_2010} 
for different surveys that have grouped these assumptions based on different criteria.
\subsection{Proposed Pipeline}
\label{sec:proposedMethod}

The proposed pipeline consists of three main components. 
Given a bag $X$  %$X {=} \{x_1,x_2,....,x_m \}$ 
of $m$ instances $x_i$, 
each of the instances $x_i$ is encoded into a feature representation $f_i$ through 
the \textit{Instance Description Unit}.
%(Section~\ref{sec:instanceDescriptionUnit}). 
Then, each element is fed to the \textit{Iterative bag Pooling Unit}, 
%(Section~\ref{sec:setPoolingUnit}), 
producing the aggregated bag representation $S$. 
Finally, a prediction $\hat{y}$ is obtained by evaluating the bag representation 
via the \textit{Prediction Unit}. % (Section~\ref{sec:predictionUnit}), 

%%%%%-------------------------------------------
\begin{figure}
\centering
\vspace{-4mm}
\includegraphics[width=1\textwidth]{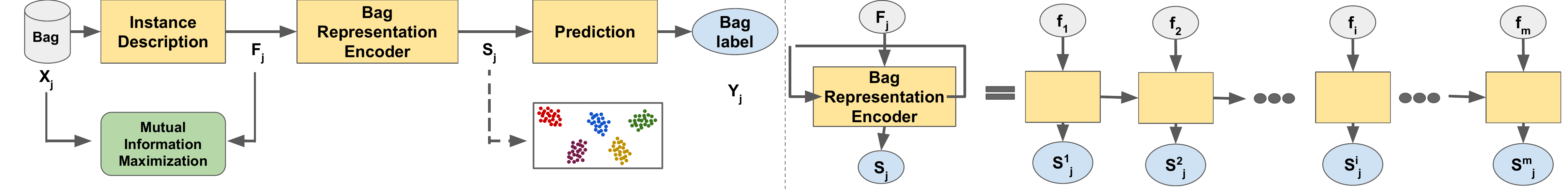}
\caption{Left: Proposed approach pipeline.
Right: Iterative bag pooling unit. 
The bag representation $S^i_j$ is updated each time the representation $f_i$ of an element is observed.}
\label{fig:proposedApproach}
% \vspace{-2mm}
\end{figure}
% \vspace{-8mm}
%%%%%-------------------------------------------

%%------------------------------
%% Instance Description Unit
%%------------------------------

\subsubsection{Instance Description Unit}
\label{sec:instanceDescriptionUnit}

This component receives the bag of instances in raw form, 
i.e. each of the instances $x_i {\in} {\mathbb{R}}^{[d]}$ 
that compose it, in its original format.
It is tasked with encoding the input bag data into a format that can 
be processed by the rest of the pipeline. 
As such, it provides the proposed method with robustness 
to different data formats/modalities.
More formally, given a dataset $\{X_j , y_j\}$ of bags $X_j$ paired with 
their corresponding bag-level labels $y_j$, each of the bags $X_j$
is encoded into a feature $F_j{=}\{f_1,f_2,...,f_m\}$. This is achieved 
by pushing each of the instances $x_i$ that compose it, through a feature 
encoder $\tau(.)$ producing the instance-level representation 
$f_i {=} \tau(x_i) , f_i \in {\mathbb{R}}^{[n]}$.
% TT: IS THIS CORRECT ? F_i is of size mxn, but f_i is only of size n, no ??

Selection of this component depends on the modality of the data to be 
processed, e.g. VGG~\cite{VGG} or ResNet~\cite{He2015DeepRL} features for still images, 
Word2Vec~\cite{MikolovWord2Vec} or BERT~\cite{devlinBERT} for text data, or 
rank-pooled features~\cite{FernandoAl:TPAMI16} or dynamic 
images~\cite{bilenDynamicImages} for video data. 
%% JO: if needed we can remove the paragraph above.

% \subsubsection{Instance-level Mutual Information Maximization}
\subsubsection{Maximizing Mutual Information from Instances}
\label{sec:mutualInf}
Mutual information can be used to 
measure the (possibly non-linear) dependency between two variables, noted as $I(A;B)$. 
Maximizing mutual information between input instance and its representation helps the model learn a better
representation~\cite{infomax2018}\cite{YangDualAEN2019}.
It is useful as a regularizer especially when learning a model from scratch.
% In a deep neural network setting, it is used in two ways: \textit{Global Mutual Information}, which measures the similarity between input instances and their global features (output of fully connected layers), and \textit{Local Mutual Information}~\cite{infomax2018} that takes into account the intermediate feature map of the input and its global features.
In our method, we follow \cite{infomax2018} where
% we maximize both global and local mutual information in order to make $f_{j}$, the global feature of input $x_j$, more representative. This can be beneficial for the following Iterative bag Pooling unit. 
% Specifically, we maximize the global and local mutual information as $I(x_{j}$;$f_{j})$ and $I(x_{j}'$;$f_{j})$, respectively.
% $x_{j}'$ is an intermediate representation of the Instance Description unit for the input instance $x_{j}$.
%
%
% The definition of MI is
% \begin{equation}
% \footnotesize
% \begin{split}
%     I(X, F) &= \iint p(x,f)log\frac{p(x,f)}{p(x)p(f)}dxdf\\
%             &= \iint p(f|x)p(x)log\frac{p(f|x)}{p(f)}dxdf
% \end{split}
% \end{equation}
%
% {where $p(x)$ is the distribution of the inputs, $p(f|x)$ is the
% distribution of the latent representations. $p(f)$ is distribution of latent space and can be calculated as $p(f) = p(f|x)p(x)$.
% $x$ can be both input instance and intermediate feature map, if it is in the first case, Eq. \ref{eq:infomax} estimates the global mutual information, if in the second case, it estimates the local mutual information
%
the total objective function is:
\vspace{-1mm}
\begin{equation}
\footnotesize
\begin{split}
    L &= \alpha \cdot max(MI_{global}) + \beta \cdot  max(MI_{local}) + \gamma \cdot Prior Matching
    \label{eq:infomax}
\end{split}
\end{equation}
% \vspace{-1mm}
$MI_{global}$ and $MI_{local}$ are the global and local Mutual information where the latter one is calculated between intermediate feature map and final representation.
$Prior Matching$ is used to match the output of the IDU to a prior: combined with the maximization of mutual information, Eq.~\ref{eq:infomax} can constrain representations according to desired statistical properties. Please refer to \cite{infomax2018} for more details regarding the derivation of Eq.~\ref{eq:infomax}.
\subsubsection{Bag Representation Encoder}
\label{sec:setPoolingUnit}
% \subsubsection{Re-visit LSTM}
The main goal of this component is to 
derive a bag-level representation $S_j {\in} {\mathbb{R}}^{[n]}$ 
that is able to encode all the instances $x_i$, and any possible 
underlying structure between them. 
As mentioned earlier, LSTM is utilized to model the underlying [unordered] structure in the bag. 

We aim at learning a bag representation that is independent of both the cardinality 
$m$ of the bag and the nature of the underlying structure. 
Starting from the element-level representations $F_j$ 
computed in the previous step, this is achieved by 
iteratively looking at the representations $f_i$, 
from each of the instances $x_i$, one at a time. 
In each iteration $i$ an updated bag-level representation 
$S^i_j$ is computed. 
In parallel, following the LSTM formulation, a feedback loop 
provides information regarding the state of the bag representation 
that will be considered at the next iteration $i{+}1$.
Finally, after observing all the $m$ instances $x_i$ in the bag, 
the final bag representation $S^m_j|_{i=m}$ is taken as 
the output $S_j$ of this component.

%% Notion
%
The notion behind this iterative bag pooling idea is that 
instances observed at specific iterations 
can be used to compute a more-informed bag-level representation 
at later iterations. Thus, allowing to encode underlying 
relationships or structures among the instances of the bag.
While this iterative assumption may hint at a sequence 
structure requirement within each bag, our empirical 
evaluation strongly suggests this not to be the case.
(see Sec. \ref{orderSelection})
%
% Moreover, this provides the proposed approach with robustness 
% towards bags possessing a sequence-like structure, while not 
% enforcing the requirement of the existence of such a 
% structure.

% In practice, this iterative mechanism can be implemented 
% through Recurrent Neural Networks~\cite{RNN_Schuster_97}, such as Long Short 
% Term Memory (LSTM) Networks~\cite{LSTM_97} or Gated Recurrent 
% Units~\cite{choGRU}, or any other machinery with means to allow
% information persistence across multiple observations $x_i$.
% %
% Here, we implement this component through LSTMs given their
% robustness of modeling structures within a bag with high 
% cardinality. This will ensure that the learned bag representation 
% can encode structures between all the elements in the set, 
% independently of the cardinality of the set.
% %
% More specifically, 

In practice, we use Bi-directional LSTMs which observe 
the instances in a bag from the left-to-right and right-to-left 
directions. This will further ensure that the context in which the 
instances of the bag occur is properly modelled.

% \vspace{-20mm}

%%------------------------------
%% Prediction Unit
%%------------------------------

\subsubsection{Prediction Unit}
\label{sec:predictionUnit}

Having a bag-level representation $S_j$ for bag $X_j$, this 
component is tasked with making a bag-level prediction 
$\hat{y_j}{=}g(S_j)$ that will serve as final output for 
the pipeline.
The selection of the prediction function $g(.)$ is related 
to the task of interest. This unit provides 
our method with flexibility to address both classification 
and regression tasks.
\section{Analysis}
\label{sec:evaluation}

%% Question-1-3 ------------------
\subsection{What kind of information can be captured by LSTMs? }
% \kaili{\subsection{How well Can LSTM handle MIL problems?}}
%% Q1: What kind of information can be captured by LSTM ? 
%% A1: info on order, but also other info - see e.g. action recognition;
%% Q3: Ok, so it can capture info that doesn't depend on order, but can it do it well ?
%% A3: Actually better than alternatives developed specificically for MIL !
%
\label{sec:MNISTexperiments}
% {Need write someting related to using LSTM to capture temporal-related feature.}
% {[CITE] }

This experiment focuses on performing multiple instance predictions based on visual 
data. Following the protocol from \cite{ITW:2018} we use images 
from the MNIST dataset~\cite{lecun-MNIST-2010} to construct image bags to define four 
scenarios, each following a different assumption: 
\textit{Single digit occurrence}, \textit{Multiple digit occurrence}, 
\textit{Single digit counting} and \textit{Outlier detection}.
%
% For the fir scenario we sample images from MNIST to construct 500 
% image bags for training and 200 bags for testing. 
% Label balance is preserved within each data split.

For this series of experiments, we use a LeNet\footnote{Please refer to the supplementary material for more details.\label{note1}}~\cite{lenet} as 
% instance descriptor unit, in order to extract image-level features; 
Instance Descriptor unit (IDU) and an LSTM with an input and cell state with 500 dimensions as Bag Representation Encoder (BRE), respectively. Both the IDU and BRE components are trained jointly from scratch.
We compare the obtained performance w.r.t. the attention-based model from \cite{ITW:2018} and the dynamic pooling method from \cite{yan18DynamicPooling}.
Mean error rate in the binary classification task is adopted
as performance metric in these experiments.

The main objective of this experiment is to answer the following questions: i) 
%\textit{\kaili{Under MI assumption}, 
whether other underlying bag structures, outside of sequential order, can be encoded properly by LSTMs?, and ii) \textit{how competitive are LTSMs when compared with methods from the MIL literature specifically designed for modeling the underlying bag structures?}

%%--------------------------------------------------------------------------

%%-------------------------------------------------------
%% Single Digit Occurrence
%%-------------------------------------------------------

\subsubsection{Single Digit Occurrence} 
In this scenario we follow the standard MI assumption and 
label a bag as positive if at least one digit '9' occurs in 
the bag. The digit '9' is selected since it can be easily 
mistaken with digit '4' and '7'~\cite{ITW:2018}, thus, 
introducing some instance-level ambiguity.
We define bags with mean cardinality $m{=}10$, and verify 
the effect that $m$ has on performance by testing two  
standard deviation values, $\sigma{=}2$ and $\sigma{=}8$.
We repeat this experiment five times generating different bags
and weight initializations. We report mean performance 
in Table~\ref{tab:MNIST} (col. II and III). %\ref{tab:singleAppearence}. 

% %%--------------------------------------------------------------------------

% \begin{table}
% \setlength{\tabcolsep}{4.7pt} % Default value: 6pt
% \centering
% \caption{Mean error rate of the Single digit occurrence.}
% % \footnotesize
% \begin{tabular*}{8.5cm}
% {@{\extracolsep{\fill}} l c c}
% \toprule 
% %
% {Method} & error ($\sigma {=} 2$) & error ($\sigma {=} 8$)   \\ 
% \midrule[0.6pt]	
% Ours    &   3.5 $\pm$ 1.1    &   \textbf{3.1$\pm$ 0.5}    \\
% Atten. Based [REF]  &   \textbf{2.8 $\pm$ 4.8}    &   4.5 $\pm$ 0.4   \\
% Gated Atten. Based [REF]    &   4.0 $\pm$ 0.9   & 4.6 $\pm$ 0.5 \\
% Dynamic Pooling [REF] &5.6 $\pm$1.1& 6.1 $\pm$ 1.2 \\
% \bottomrule[1pt]
% \end{tabular*}
% \label{tab:singleAppearence}
% % \vspace{-6mm}
% \end{table}
% %%--------------------------------------------------------------------------

\textbf{Discussion:} The results indicate that, in this
task, our performance is comparable with the state-of-the-art 
for lower values of $\sigma$ and superior as $\sigma$ increases.
This is to some extent expected, since at lower $\sigma$ the 
cardinality (i.e. the number of instances) of each bag is almost 
fixed. This setting is favorable for the attention-based method 
since it operates in a feed-forward fashion. Yet, note 
the high standard deviation in performance produced by this baseline.
On the contrary, at higher $\sigma$ values there is a higher variation 
of cardinality across bags. 
%As can be noted in Table~\ref{tab:MNIST} (col. III) %\ref{tab:singleAppearence} 
Under this setting, feed-forward approaches start to produce higher 
errors. Here our method produces superior performance, ${\sim}1.4$ 
percentage points (pp) w.r.t. to the state-of-the-art.

%%--------------------------------------------------------------------------
\vspace{-6mm}
\begin{table*}
\setlength{\tabcolsep}{4.7pt} % Default value: 6pt
% \vspace{-4mm}
\centering
\caption{Mean error rate (in percentage points) of experiments considering digits from the MNIST dataset. (*) refers to baselines which include the Mutual Information loss.}
% \footnotesize
\resizebox{1\columnwidth}{!}{%
\begin{tabular*}{18cm}
{@{\extracolsep{\fill}} l c c c c c c}
\toprule 
% Method & Try-on ROI (SSIM/LPSIS-VGG) & Take off (SSIM/LPSIS-VGG)  \\
{Method} & single digit($\sigma {=} 2$) & single digit($\sigma {=} 8$) & multiple digits & digit counting  & outlier detection \\ 
\midrule[0.6pt]	
Atten. Based  &   \textbf{2.8 $\pm$ 4.8}    &   4.5 $\pm$ 0.4    & 28.5 $\pm$ 0.7   &33.4 $\pm$ 19.3& 37.0*   \\
Gated Atten. Based    &   4.0 $\pm$ 0.9   & 4.6 $\pm$ 0.5 & 27.4 $\pm$ 0.9   &11.9 $\pm$ 3.6   & 37.4*  \\
Dyn. Pool &5.6 $\pm$1.1& 6.1 $\pm$ 1.2 & 28.5 $\pm$ 6.6   &25.4 $\pm$ 1.8& 40.9* \\
Ours w/o Mut. Info.    &   3.5 $\pm$ 1.1    &   \textbf{3.1$\pm$ 0.5} & 6.4$\pm$ 1.4 &9.0$\pm$ 2.7  & 50.0  \\
Ours & 4.0$\pm$ 0.4 &4.1$\pm$1.4& \textbf{3.5$\pm$1.3}&\textbf{7.4$\pm$1.2}&\textbf{2.07}\\
\bottomrule[1pt]
\end{tabular*}
}
\label{tab:MNIST}
\vspace{-2mm}
\end{table*}
\vspace{-13mm}
%%-------------------------------------------------------
%% Multiple Digit Occurrence
%%-------------------------------------------------------
% \vspace{-5mm}

\subsubsection{Multiple Digit Occurrence} This is an extension 
of the previous scenario in which instead of focusing on the 
occurrence of a single digit class, the model should recognize 
the occurrence of instances of two digit classes \kaili{(presence-based MI assumption~\cite{foulds_frank_2010})}. 
More specifically, a bag is labeled positive if both digits '3' 
and '6' occur in it, without considering the order of occurrence. 
For this scenario 1,000 bags are sampled for training. 
Results are reported in Table~\ref{tab:MNIST} (col. IV).

% %%--------------------------------------------------------------------------
% \begin{table}
% \setlength{\tabcolsep}{4.7pt} % Default value: 6pt
% \centering
% \caption{Mean error rate of the multiple digit occurrence.}
% % \footnotesize
% \begin{tabular*}{8.5cm}
% {@{\extracolsep{\fill}} l c}
% \toprule 
% % Method & Try-on ROI (SSIM/LPSIS-VGG) & Take off (SSIM/LPSIS-VGG)  \\
% {Method} & error rate   \\ 
% \midrule[0.6pt]	
% Ours & \textbf{0.064$\pm$ 0.014}    \\
% Atten. Based & 0.285 $\pm$ 0.007\\
% Gated Atten. Based & 0.274 $\pm$ 0.009\\
% Dynamic Pooling. Based & 0.285 $\pm$ 0.066\\
% \bottomrule[1pt]
% \end{tabular*}
% \label{tab:multiAppearence}
% % \vspace{-6mm}
% \end{table}
% %%--------------------------------------------------------------------------

\textbf{Discussion:}
It is remarkable that when making this simple extension of considering 
the occurrence of multiple digits, i.e. '3' and '6', the state-of-the-art 
methods suffer a significant drop in performance. 
This drop put the state-of-the-art methods ${\sim}27$ pp below, on average,
w.r.t. the performance of our method. 
Please note that in this experiment the order (or location) of the 
two digits does \textit{not} matter. \tocheck{This supports previous observations that LSTMs can indeed handle 
multiple instances of interest, independent of the ordering in which 
they occur within the bags.} 
%
% Compared to the \textit{Single digit occurrence} 
In this scenario, 
where observing multiple instances is of interest, the model needs to 
``remember'' the information that it has seen in order to asses whether 
instances of the classes of interest have been encountered. 
The feed-forward models lack information 
persistence mechanisms; which translates to a poor ability to 
remember and to handle multiple instances of interest.
Surprisingly, in spite of its iterative nature, the Dynamic pooling method 
is not able to preserve the information it has observed across iterations, 
resulting in similar performance as the other baselines.

%%-------------------------------------------------------
%% Digit Sequences
%%-------------------------------------------------------

% \subsubsection{Digit Sequences} Similar to the previous setting, 
% in this scenario multiple elements are of interest within each 
% set, however, the order of occurrence of these \textit{do} matter.
% More specifically, a bag is labeled positive if an instance of 
% digit '3' occurs earlier, i.e. it has a lower $i$ index in 
% the set, than one of digit '6'.
% This scenario follows the Rank-based MI assumption presented in 
% Section~\ref{sec:MIAssumptions}.
% Quantitative results are reported in Table~\ref{tab:MNIST} (col. V).
% %\ref{tab:squentialNumber}.

% \textbf{Discussion:}
% As can be seen Table~\ref{tab:MNIST}, under this 
% scenario, the proposed method leads the performance table 
% by a large margin of ${\sim}45$ pp. This is to some extent 
% expected since the LSTM network used to implement our 
% iterative bag pooling unit is designed to handle bags whose 
% instances posses an underlying sequential structure.
% % % % Moreover, this hints at the possibility of applying the 
% % % % proposed method for problems where there rank-based assumption 
% % % % holds, e.g. action recognition, language modeling, etc.

%%-------------------------------------------------------
%% Digit Counting
%%-------------------------------------------------------

\subsubsection{Digit Counting} 
Previous scenarios addressed the classification task of predicting 
positive/negative bag-level labels. In contrast, in this scenario, 
we focus on the regression task of counting the number of 
instances of a specific digit class of interest within the bag \jose{(presence-based MI assumption)}.
In order to make our approach suitable to address a regression problem, 
instead of using a classifier as prediction unit we use a regressor 
whose continuous output is rounded in order to provide a 
discrete count value as output.
In this experiment the digit '9' is selected as the class to be counted. 
The mean cardinality of each bag is fixed to $m{=}15$.
Performance is reported in Table~\ref{tab:MNIST} (col. V).
% in terms of mean error rate.

\textbf{Discussion:}
%A quick inspection to 
From Table~\ref{tab:MNIST} (col. V) 
%reveals that, in this experiment, its is still present 
the same trend can be observed: our method has 
superior performance and higher stability than the attention-based model and other baselines.
When conducting this counting task, our method obtains a performance 
that is superior by $24$ pp w.r.t. the attention-based model and by 
$16$ pp w.r.t. the dynamic pooling.
{These results further confirm the capability of LSTMs to handle this type of unordered regression problems \cite{LSTM_counting_19}.}

%%-------------------------------------------------------
%% Digit Outlier Detection
%%-------------------------------------------------------
\vspace{-2mm}
\subsubsection{Digit outlier detection}
This task is concerned with identifying whether a bag contains a digit which is
different from the majority (outlier). Different from \textit{Single digit occurrence}, this task is more difficult since the model has to understand:  
% 1. the majority digits, 2. the outlier digit, while in \textit{Single digit occurrence} it only needs to find the specific digits. 
i) the two digit classes that might be present in the bag, and ii) the proportion condition that makes the bag an outlier. 
This is different from \textit{Single digit occurrence} where it only needed to identify the "witness" digit '9'.
Besides, there is no restriction on the outlier and majority digits, they can be any digit class from MNIST dataset.
\kaili{This constitutes a collective MI assumption since all the instances determine the underlying structure of the bag.}
Therefore, given the complexity of this task, in this experiment we apply the mutual information loss on every baseline method in order to assist their training.
We use 10,000 bags to train the model and 2,000 bags to test. The bag cardinality is 6 with 1 standard deviation. Table \ref{tab:MNIST} (col. VI) shows quantitative results of this experiment. 

\textbf{Discussion:}
It is remarkable that, even after applying the mutual information loss on the other baselines, they still have a low performance on this task. We notice that the Attention and Gated Attention methods work slightly better than Dynamic Pooling. More importantly, our method, based on LSTMs, outperforms the baselines by a large margin (${\sim}36$ pp).
% {By maximizing the mutual information the IDU can learn better instance representations. This seems to help the LSTMs 
% address the two challenges mentioned above.
% % understand the occurring digit classes and the difference in proportion of occurrence.
% }
This suggests that LSTMs are quite capable at modeling this type of bag structure, even to the point of outperforming MIL methods tailored to model bag-based structures.

% {Because the ISP unit in our model has memorize ability, helping the model to distinguish the majority digits and the outlier in a set.} 

%% Question-2 ------------------
\subsection{Does the result depend on the order chosen?}
\label{orderSelection}
%% Q2: But how do I select the order if I'm given an orderless bag ? Does the result depend on the order chosen ?
%% A2: Show results for different orders for MNIST experiments.
% {Describe experiments where instances are fed in different orders to the LSTM}

The short answer is no. The reason is that in the training phase we push the bags with different orders (as a form of data augmentation) to the model while the bag labels are the same. By following this procedure the loss function will not penalize differences in the order in which the instances are observed.
To further verify this, we repeated the test phase of our experiments 100 times with the contents of each bag (cardinality=m, in total m! combinations) shuffled thus producing bags with 100 different orders. 
% Then we measure the rate of change
% w.r.t the most frequent prediction obtained from the 100 tests 
% and report the mean performance over all the sets. 
% The obtained rate of change is 
% % 0.21+- 1.39\%, 0.13 +- 0.81\%, 2.81+- 7.35\% 
% (0.21$\pm$1.39)\%, (0.13$\pm$0.81)\%, (2.81$\pm$7.35)\%
\kaili{Then, similar to Sec. \ref{sec:MNISTexperiments}, we measure the error rate and report the mean performance.
The obtained error rate is 
(4.2$\pm$0.6)\%, (3.5$\pm$0.4)\%, (7.8$\pm$0.7)\%
for the Single-digit, Multiple-digit and Digit counting experiments, respectively ($m{=}10, 12, 15$ respectively). They are very close to the numbers reported in Table \ref{tab:MNIST}. }
The results \kaili{verify} that the LSTM is able to learn, to a good extent, that the underlying MIL assumptions were permutation invariant - changing the order of instances of a bag has a relatively low effect on the 
% classification/regression 
prediction in most of the cases.
% \kaili{"copy from RBT": LSTMs are not completely order-independent, they are pretty robust to it. They can outperform the baseline methods which is total permutation invariant.}

%% Question-2.5 ------------------
\subsection{Does the result depend on the cardinality of the bag?}
%% Q2: But how do I select the order if I'm given an orderless bag ? Does the result depend on the order chosen ?
%% A2: Show results for different orders for MNIST experiments.
% No, the model we propose is robust enough to bags of various cardinalities (sizes).
No, modeling bag representations via LSTMs seems robust enough to bags with different cardinalitiy (sizes).
We verify this by conducting an extended experiment based on the \textit{multiple instance occurrence} scenario. 
Firstly, we consider bags with higher cardinality but keep \textit{only one relevant instance pair (‘3’,‘6’) present} using one of our trained models (mean bag cardinality $m{=}12$). 
We obtained error rates of 7\%, 14.5\%, 42\%, and 44\% for bag cardinality 20, 50, 100 and 200, respectively. 
This result is not surprising since during training the bag cardinality was much lower. To have a fair experiment, we use bags with higher cardinality to finetune our model, using 1/5 amount of the original number of training bags (i.e. now we use 200 bags).
Similarly, the larger bags still contain only one pair of relevant digits.
This results in error rates of (2.38$\pm$0.41)\%, (3.13$\pm$0.89)\%, (4.25$\pm$1.3)\% 
for mean bag cardinality of 50, 100 and 200, respectively.
This shows that LSTMs are still capable of modeling unordered bags even when bags with significantly higher cardinality are considered, although, unsurprisingly, training and testing conditions should match.

\subsection{Effect of the Complexity of the Data}
%% Q4: Ok, but does this work beyond MNIST, on some more real-world problems ?
%% A4: Sure, here's results on common benchmark datasets used in the MIL literature
In this section, we shift our analysis to real-world data.
Summarizing, the results show that our method still works comparable and even better than the baselines.

%%------------------------------
%%  Cross-domain Retrieval
%%------------------------------
%%%%%-------------------------------------------
\vspace{-7mm}
\begin{figure}
\centering
\includegraphics[width=1\textwidth]{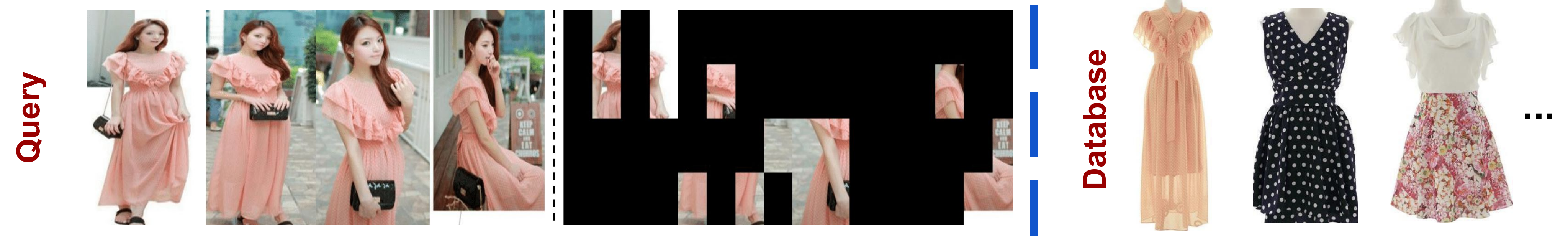}
\caption{Examples of instances for the original (left), occluded (middle) and database images (right) in our cross-domain clothing retrieval experiment.
% {might be good to add the unoccluded version of this figure just above it}
}
\label{fig:exampleLookbook}
\vspace{-2mm}
\end{figure}
\vspace{-14mm}

%%%%%-------------------------------------------

\subsubsection{Cross-domain clothing retrieval}
For this experiment, we divide images from the Lookbook dataset into 
two domains: catalog clothing images and their corresponding human 
model images where a person is wearing the clothing product, see Fig. \ref{fig:exampleLookbook}. 
Each clothing product has one catalog image and several human
model images. We only consider the products with 
five or more human model images, resulting in 6616 unique products 
(latent classes $c_i$) with around 63k images in total.
Every product image has 5-55 human model images.
The training bag contains 4000 classes while the validation and test bags 
have 616 and 2000 classes, respectively. We run two experiments on
this dataset as described in the following sections.
Given the higher complexity of images in this dataset, we use a pre-trained
VGG16\footnote{Please refer to the supplementary material for more details.\label{note2}}~\cite{VGG}
as IDU.
Since this unit is pretrained, the mutual information loss is not applied for this unit in this experiment.
Moreover, 
%for the iterative bag pooling unit, 
we set the dimensionality of the input and cell state 
of our LSTM to $n{=}2048$. 
% {As we mentioned in Sec.\ref{sec:instanceDescriptionUnit}, the mutual information loss is not applied in this experiment.}
% \vspace{-1mm}

For this experiment, human model images are used as queries while 
catalog images serve as database, 
thus, defining a many-to-one retrieval setting. 
% TT: MANY-TO-ONE OR ONE-TO-MANY ? ABOVE YOU SAID THERE WAS ONLY
% ONE CATALOG IMAGE FOR EACH PRODUCT ?
The cardinality of each bag is the same as the number of 
human model images of each product (class).
We conduct two variants of this experiment. On the first variant 
we use the complete image, as it is originally provided.
%the traditional image retrieval setting, 
The second is an occluded variant where every human model image in 
a bag is divided into a $4{\times}4$ grid of 16 blocks. 
12 of these blocks are occluded by setting all the pixels therein to 
black. 
By doing so, every single image in a bag can only show part of the 
information while their combination (i.e. the whole bag) represents 
the complete clothing item. 
Catalog images in the database are not occluded in this experiment. 
\kaili{This experiment can be regarded as an extreme case of standard MI assumption, where all the instances in each bag is positive.}
% The catalog images are not occluded since every product has only one 
% catalog image.

%
As baselines, in addition to the attention-based model
%~\cite{LiuCVPR17attention}, 
we follow DeepFashion~\cite{DeepFashion}, and train a model to perform 
retrieval by computing the distances by considering single image 
representations instead of bag-based representations. 
Following the multiple queries approach from \cite{multipleQueries},
we report performance of three variants of this method: \textit{Single-AVE}, 
where the distance of each bag is computed as the average of the distances from 
every image in the bag w.r.t. an item in the database; 
\textit{Single-MIN}, where the distance of the bag is defined as the minimum 
distance of an image in the bag w.r.t. an item in the database; and
\textit{Single Fea. AVE }, where the distance of the bag is calculated 
as the distance of a prototype element w.r.t. an item in the database. 
As prototype element 
%is computed by computing 
we use the average feature 
representation $\overline{f_i}$ from the representation $f_i$ of every 
element in the bag. We refer to these baselines as \textit{Single-image models}.

% This retrieval task is to some extent related to the collective MI 
% assumption~(Sec.~\ref{sec:MIAssumptions}) since all the elements in the bag 
% contribute to the task handled by the model.

%%--------------------------------------------------------------------------
\hskip -16pt
\begin{minipage}{\textwidth}
% \vspace{-6mm}
\begin{minipage}[b]{0.48\textwidth}\setlength{\tabcolsep}{4.7pt} % Default value: 6pt
% \parbox{.45\linewidth}{
\centering
% \caption{Retrieval on the original Lookbook dataset.}
% \footnotesize
\resizebox{1\columnwidth}{!}{
\begin{tabular*}{9cm}
{@{\extracolsep{\fill}} l c c c c }
\toprule 
{Method} & rec.@1 & rec.@10 & rec.@20 & rec.@50  \\ 
\midrule[0.6pt]	
Atten. & 13.75& 39.25& 49.70 &63.60\\
Dyn. Pool & 16.75& 47.65& 59.45 & 73.60\\
Single AVE &20.55 & 57.05& 68.25& 81.90\\
Single MIN &22.60 & 58.15& \textbf{69.20} & 82.50\\
Single Fea. AVE &20.15 & 56.25& 67.85 & 81.50\\
Ours w/o mut. info. &\textbf{22.95} &\textbf{58.65}& 68.70& \textbf{83.00}    \\
\bottomrule[1pt]
\end{tabular*}}
\vspace{-3mm}
\captionof{table}{Retrieval on the original Lookbook dataset.}
\label{tab:fullImageRetrieval}
\end{minipage}
\hskip 10pt
\begin{minipage}[b]{0.48\textwidth}
\centering
% \footnotesize
\resizebox{1\columnwidth}{!}{
\begin{tabular*}{8.5cm}
{@{\extracolsep{\fill}} l c c c c }
\toprule 
{Method} & rec.@1 & rec.@10 & rec.@20 & rec.@50  \\ 
\midrule[0.6pt]	
Atten. & 3.55 & 20.6 & 32.95 & 53.65\\
Dyn. Pool & 1.95 & 11.95 & 29.35 & 32.55\\
Single AVE & 3.65& 23.85& 35.06&56.10\\
Single MIN & 5.25& 26.05&37.35&55.00\\
Single Fea. AVE & 5.10& 25.60&36.95&54.65\\
Ours w/o mut. info. & \textbf{9.25} & \textbf{34.75}& \textbf{45.00}& \textbf{61.80}    \\
\bottomrule[1pt]
\end{tabular*}}
\vspace{-3mm}
\captionof{table}{Retrieval on the occluded Lookbook dataset.}
\label{tab:occluImageRetrieval}
\end{minipage}
\end{minipage}
% \vspace{-6mm}
% \end{table}
%%--------------------------------------------------------------------------

\textbf{Discussion:}
Table~\ref{tab:fullImageRetrieval} shows that in the original 
setting our method tends to obtain superior recall values in 
the majority of the cases, with the exception of the case when 
the closest 20 items (recall@20) are considered.
When looking at the occluded variant of the 
experiment, a quick glance at Table~\ref{tab:occluImageRetrieval} 
shows that, compared to the original setting, absolute 
performance values on this setting are much lower.
This is to be expected since this is a more challenging scenario 
where the model needs to learn the information cumulatively 
by aggregating information from parts of different images.
In this occluded setting, our method clearly outperforms 
all the baselines.
% the attention-based, 
% %~\cite{LiuCVPR17attention}, 
% dynamic pooling
% %~\cite{yan18DynamicPooling} 
% and the methods based on single-image distances. 
%
This could be attributed to the information persistence component 
from the LSTMs. This component allows our method to 
select what to remember and what to ignore from each of the instances 
that it observes when updating the bag representation used to compute 
distances.
The difference w.r.t. to the 
\textit{Single-AVE} and  \textit{Single-MIN}
% \textit{Single-AVE},  \textit{Single-MIN} and \textit{Single Fea. AVE}
%\textit{Single-image model}
baselines is quite remarkable given that they require a significant 
larger number of element-wise distance computations w.r.t. items in the 
database.
This may lead to scalability issues when the dataset size increases, 
as the computation cost will grow exponentially.

Moreover, in both occluded and non-occluded datasets, 
we notice that the 
% \textit{Single-AVE}, \textit{Single-MIN} and \textit{Single Fea. AVE} 
\textit{Single-image model}
baselines 
have a superior performance w.r.t. the attention-based model
%~\cite{LiuCVPR17attention} 
and dynamic pooling model. 
%~\cite{yan18DynamicPooling}.
We hypothesize that is because  the single-image models can better exploit 
important features, e.g. discriminative visual patches, since they compute 
distances directly in an instance-wise fashion. 
In contrast, it is likely that some of these nuances might get averaged 
out by the feature aggregation step that is present in the 
attention-based model.
%~\cite{LiuCVPR17attention}. 

% %%%%%-------------------------------------------
% \begin{figure*}[h]
% \centering
% \includegraphics[width=1\textwidth]{imgs/interpolation_v3-crop.pdf}
% \caption{t-SNE visualization of the learned bag representation. 
% The first two rows show examples of predictions on true positive and true negative sets, 
% except for the \textit{Digit Counting} experiment, which shows two bags containing 4 and 0, 
% elements of interest, respectively. 
% The third row shows the prediction of 20 examples overlaid on the t-SNE space for the 
% digit-based experiments.}
% \label{fig:modelExplanation}
% \end{figure*}
% %%%%%-------------------------------------------

%%------------------------------
%%  Colon Cancer Prediction
%%------------------------------

\vspace{-6mm}
\subsubsection{Colon Cancer Prediction}
\label{sec:colonExp}
 % {Detecting the cancerous region from hematoxylin and eosin (H\&E) stained whole-slide images is an interesting and important task in the field of clinical medicine. However, annotating the pixel level labels costs a lot of time. 
% One of the solutions is to work with weak labels\cite{ITW:2018}, i.e. label of one full H\&E image.}
This task consists of predicting the occurrence of Colon cancer from histopathology images.
The used Colon cancer dataset contains 100 $500{\times}500$ H\&E images with a total of 22k annotated nuclei. 
There are four types of nuclei: \textit{epithelial, inflammatory, fibroblast}, and \textit{miscellaneous}.
This experiment focuses on identifying whether colon cancer histopathology images contain a specific type of nuclei. %which indicates the presence of cancer.
We follow the protocol from \cite{ITW:2018} and treat every H\&E image as a bag composed by instances (patches) of $27{\times}27$ pixels centered on detected nuclei. The bag cardinality varies from 6 to 796 depending on the number of nuclei present in the image.
\kaili{Following a standard MI assumption,} a bag is considered positive if it contains \textit{epithelial} nuclei since Colon cancer originates from epithelial cells~\cite{ITW:2018}\cite{Ricci-Vitiani2007} 
This produces a dataset with 51 and 48 positive and negative bag examples, respectively.
%51 H\&E images contain the epithelial nuclei while 48 of them do not.
We extend this dataset via data augmentation as in \cite{ITW:2018}. 
%In order to avoid the overfitting, we follow\cite{ITW:2018} to do a similar data argumentation. 

We adapt an architecture which is similar to \cite{colon_cancer_2016} to define the IDU and a 512 dimension input and cell state to define the LSTM (BRE). The whole model is trained from scratch. Following the protocol, only bag-level binary labels are used for training.
We conduct experiments considering the same baselines as in previous experiments. We apply five-fold cross validation and report the mean performance and standard deviation. For reference, we also provide the results presented in \cite{ITW:2018} for the baselines Atten.* and Gated Atten.*.
Table~\ref{tab:colonCancerExp} shows quantitative results in terms of Accuracy and F1-Score.  

\textbf{Discussion:} 
This experiment, where a bag can have up to 796 instances, serves a good test-bed to assess the performance of the proposed method on bags with high cardinality.
%
%On the satifactory side, f
From the results in Table~\ref{tab:colonCancerExp}, we can notice
that our method still outperforms all the considered baselines. 
\vspace{-6mm}
\begin{table*}
\setlength{\tabcolsep}{4.7pt} % Default value: 6pt
% \vspace{-6mm}
\centering
\caption{Colon cancer experiment results.}
% \footnotesize
\scalebox{0.7}{%
\begin{tabular}{l c c }
\toprule 
% Method & Try-on ROI (SSIM/LPSIS-VGG) & Take off (SSIM/LPSIS-VGG)  \\
{Method} & Accuracy  & F1-Score    \\ 
\midrule[0.65pt]	
Atten.* & 90.40$\pm$1.10 & 90.10$\pm$1.10\\
Gated Atten.* & 89.80$\pm$2.00 &  89.30$\pm$2.20\\
\midrule[0.6pt]	
Atten. & 88.79$\pm$6.16 & 88.85$\pm$6.35\\
Gated Atten. & 86.89$\pm$3.93 &  86.87$\pm$6.67\\
Dyn. Pool & 87.89$\pm$2.37 &  88.18$\pm$2.11\\
Ours w/o mut. info. & 90.89$\pm$2.06  & 90.66$\pm$2.80\\
Ours & \textbf{92.74$\pm$2.41}  & \textbf{93.08$\pm$1.36}\\

\bottomrule[1pt]
\end{tabular}
}
\label{tab:colonCancerExp}
\vspace{-2mm}
\end{table*}
\vspace{-4mm}
\section{Internal State Inspection and its application}
%% Q5: Can I learn something from observing the internal state of the LSTM during processing ?
%% A5: Yes, it gives nice insight in the internal 'reasoning' + experiment
% {I suggest this section }
% The above experiments clearly show that LSTM model can capture the instance occurrence information and accumulate the information of each instance in a bag without taking into account the instance order and it works comparable and even better than the traditional methods.
% LSTM can capture the temporal information of a set, therefore, it can predict the future instance [CITE].
Previous efforts~\cite{LSTM_97,LiangICCV17,LSTM_CV_15} based on ordered bags have shown that the internal state of the representation within the LSTMs can be used to predict future instances.
Here, we have shown that LSTMs can also encode other types of bag-based information internally. This begs the question - \textit{what else can the internal representation in LSTMs reveal in the unordered setting?}. 
Here we conduct an analysis aiming to answer this question.
% To answer this question, here make an inspection the internal states changes of LSTMs as it observes instances.
\vspace{-4mm}
\subsection{Internal States Inspection}
\begin{figure}[h]
\centering
\includegraphics[width=1\textwidth]{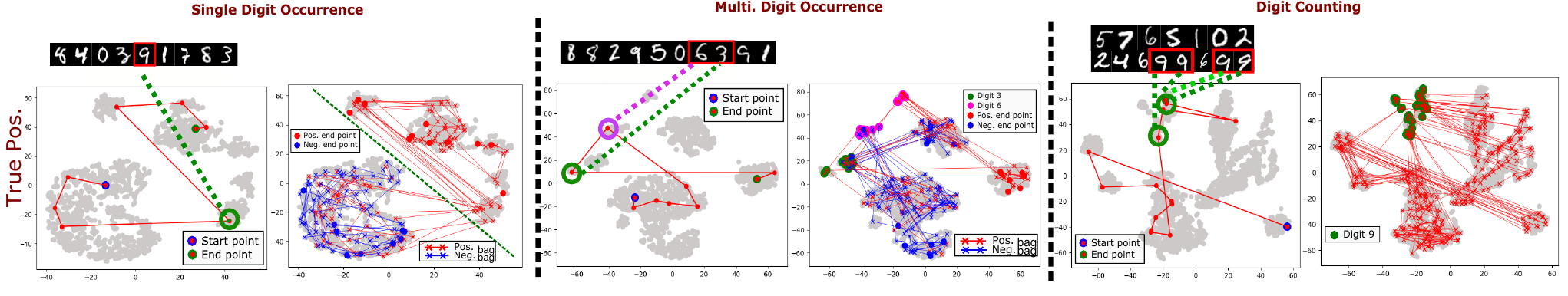}
\caption{t-SNE visualization of the interal states for three MNIST-based experiments. The left figures show an example of predictions on true positive bag except for the Digit Counting experiment which shows a bag containing 4 instances of interest. The right figures show the prediction of 20 examples overlaid on the t-SNE space for the digit-based experiments. The red and blue lines refer to positive and negative bags, respectively. Best viewed in color.}
\label{fig:internalStateInspection}
\vspace{-2mm}
\end{figure}

% We observe how LSTMs' internal states change by pushing bags to our model, collecting all the internal states (including the final one) and using t-SNE to visualize them.
%
We begin by making an inspection of the internal states of the LSTM as it observes instances and use t-SNE~\cite{tsne} to visualize these internal states.
Fig \ref{fig:internalStateInspection} clearly shows that when the model meets (one of) the condition(s), the current internal state changes evidently, which means LSTMs can distinguish the condition instances with other instances within the bags.
Based on this observation, we wonder whether it is possible to predict the instance level label based on the bag level label. 

\vspace{-8mm}
%%====================================
%%====================================
\vspace{-12mm}

% In this section, we investigate more on this property of LSTM model.
\subsection{Weakly Supervised Learning of Instance-Level Distributions}

\label{sec:weaklySupervise}
We have presented using LSTM 
to make predictions from a bag-level representation $S_j$ 
through the use of a prediction function $g(\cdot)$. 
There is a connection between the 
% {\textit{Standard MI assumption}} 
MIL task and the distribution of the instance representation. 
% For example, in order to manage the \textit{Standard MI assumption} task, the model should be able to distinguish between the "witness" instance, which satisfies the underlying MI property $\alpha$, and other instances. Here we put two hypotheses:{I am not sure about if we can write like this, but it looks good.}
Based on this observation we put forward the following hypotheses:
%
% Therefore, two interesting questions come: 
% can the model trained for MIL task model the instance distribution? 
% Further more, can we predict the instance label based on it?

\textit{- Hypothesis 1: A model trained for a MIL task can learn the underlying distribution over the instances.}
% \textit{Hypothesis 2: The model that is trained for a MIL task can be used to predict instance labels if it is a binary task and needs identify two factors of the instances.}

\textit{- Hypothesis 2: A prediction function $g(\cdot)$ trained on the bag representation $S$ can be used to make instance-level predictions if the distribution from $S$, influenced by the underlying MI assumption, is close to that of $F$.}

We propose the following approach to recover the underlying instance-level representation and make instance-level predictions. %
We break down the instance bag $X{=}\{x_1, x_2, ..., x_m\}$ into $m$ singleton bags $X_{1} {=} \{x_1\}$, $X_{2} {=} \{x_2\}$, ..., $X_{m} {=} \{x_m\}$. The singleton bag $X_{j} {=} \{x_j\}$ is sent to the model, passing the IDU and the LSTM.
Afterwards, the output $S_j$ of the LSTM from every singleton is collected into a feature matrix $\mathbb{S}, \mathbb{S}{\in} \mathbb{R}^{[m {\times} n]}$. 
Then, k-means clustering algorithm is applied on $\mathbb{S}$ with the number of clusters determined by the corresponding MIL task.
We use a similar metric to clustering purity, where we calculate the purity of each cluster first and average them instead of calculating the purity of all samples. By doing this we avoid problems caused by imbalanced data.
The clustering performance  reflects the ability of modeling the distribution of instances for the model.
\vspace{-3mm}

\subsection{Weakly Supervised Instance-level Learning}
In Sec.~\ref{sec:weaklySupervise} we presented two hypotheses related to the weakly supervised instance-level learning. We will address them in this section.

\textbf{Modelling Instance-level Representations}
% \subsubsection{Modelling Instance-level Representations}
In Sec.~\ref{sec:MNISTexperiments} and \ref{sec:colonExp}, we trained both IDU and LSTM from scratch by considering the bag-level labels only. This can be regarded as weakly supervised learning if the goal is to make instance-level predictions.
% To evaluate the model, we treat every instance as a singleton set.
% We collect the output of the IDU for every singleton bag and run KMeans clustering algorithm on top of it. 
For attention-based methods, we collect the output of IDU and multiply with weight 1, since it is a singleton bag and there is no LSTM.
Following this procedure, both methods use the features after their respective units handling the MIL task.
We evaluate instances from both testing/training bag for the baseline and our model, respectively.
We choose the Gated-Attention model as a baseline since it works best among the attention-based methods in Sec.~\ref{sec:MNISTexperiments}.
Table \ref{tab:instanceLabelMNISTExp} reports the clustering performance metric described in Sec.~\ref{sec:weaklySupervise}.

%%--------------------------------------------------------------------------
% \begin{table}
% \begin{minipage}{\textwidth}
% \hskip -15pt
% \begin{minipage}[b]{0.35\textwidth}
% \setlength{\tabcolsep}{4.6pt} % Default value: 6pt
% \centering
% % \footnotesize
% \resizebox{1\columnwidth}{!}{
% \begin{tabular*}{9cm}
% {@{\extracolsep{\fill}} l c c }
% \toprule 
% \multirow{2}{*}{Task} & Gated Atten.  & Ours  \\ 
%                     & (test/train) &(test/train)\\
% \midrule[0.6pt]	
% \textit{single digit} (2 classes) & 98.69/98.92&97.59/97.42 \\
% \textit{multiple digits} (3 classes) &   85.92/87.47 & \textbf{97.94}/\textbf{97.06} \\
% \textit{digit counting} (2 classes) &99.22/99.31&99.15/99.23\\
% \textit{outlier detection} (10 classes) &59.33/57.02&\textbf{97.96}/\textbf{97.52}\\
% \bottomrule[1pt]
% \end{tabular*}
% }
% \captionof{table}{Instance clustering accuracy from MNIST-bag task models.}
% \label{tab:instanceLabelMNISTExp}
% \end{minipage}
% % \vspace{-2mm}
% % \end{table}
% % %%--------------------------------------------------------------------------
% % \begin{figure}
% \hskip 5pt
% \begin{minipage}[b]{0.65\textwidth}
% \centering
% \includegraphics[width=1\textwidth]{imgs/tsne_mnist_bag_v10.pdf}
% \captionof{figure}{t-SNE visualization of features extracted from \textbf{our} MIL model in three MNIST bag tasks (left) and with baseline models in \textit{Outlier detection} (right). }
% \label{fig:tsneMNIST}
% % \vspace{-6mm}
% % \end{figure}
% \end{minipage}
% \end{minipage}

\vspace{+2mm}
\begin{minipage}{\textwidth}
\hskip -15pt
\begin{minipage}[b]{0.45\textwidth}
\setlength{\tabcolsep}{4.6pt} % Default value: 6pt
\centering
% \footnotesize
\resizebox{1\columnwidth}{!}{
\begin{tabular*}{9cm}
{@{\extracolsep{\fill}} l c c }
\toprule 
\multirow{2}{*}{Task} & Gated Atten.  & Ours  \\ 
                    & (test/train) &(test/train)\\
\midrule[0.6pt]	
\textit{single digit} (2 classes) & 98.69/98.92&97.59/97.42 \\
\textit{multiple digits} (3 classes) &   85.92/87.47 & \textbf{97.94}/\textbf{97.06} \\
\textit{digit counting} (2 classes) &99.22/99.31&99.15/99.23\\
\textit{outlier detection} (10 classes) &59.33/57.02&\textbf{97.96}/\textbf{97.52}\\
\bottomrule[1pt]
\end{tabular*}
}
\vspace{-2mm}
\captionof{table}{Instance clustering accuracy from MNIST-bag task models.}
\label{tab:instanceLabelMNISTExp}
\end{minipage}
% \vspace{-2mm}
% \end{table}
% %%--------------------------------------------------------------------------
% \begin{figure}
\hskip 5pt
\begin{minipage}[b]{0.55\textwidth}
\centering
% \caption{Instance label accuracy for Colon cancer dataset}
% \footnotesize
\resizebox{1\columnwidth}{!}{
\begin{tabular*}{10.5cm}
{@{\extracolsep{\fill}} l c c c }
\toprule 
\multirow{2}{*}{Method} & TP  & TN & mean Acc \\ 
&(test/train) & (test/train) & (test/train)\\
\midrule[0.6pt]	
Atten. & 32.42~/~21.25 &\textbf{98.45~/~99.22}& 65.43~/~63.60\\
Ours & \textbf{73.47~/~70.73} & 92.39~/~92.28 &\textbf{82.93~/~81.51}\\
Supervised & \textit{78.92~/~92.09} &\textit{91.14~/~98.22}& \textit{85.03~/~95.16}\\
\bottomrule[1pt]
\end{tabular*}
}
\vspace{-2mm}
\captionof{table}{Instance label accuracy for Colon cancer dataset.}
\label{tab:instanceLabelcolonCancerExp}
\end{minipage}
\end{minipage}
% \vspace{-8mm}

\vspace{-5mm}
\begin{figure}
\centering
\includegraphics[width=1\textwidth]{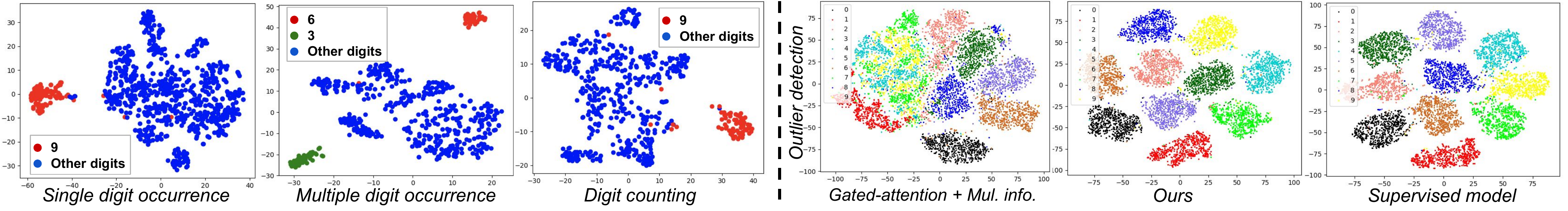}
\caption{t-SNE visualization of features extracted from \textbf{our} MIL model in three MNIST bag tasks (left) and with baseline models in \textit{Outlier detection} (right).}
\label{fig:tsneMNIST}
\vspace{-2mm}
\end{figure}
\vspace{-6mm}

\textbf{Discussion:}
Table \ref{tab:instanceLabelMNISTExp} indicates that for simple tasks, such as \textit{single digit occurrence} and \textit{digit counting}, both attention-based and our methods can distinguish the background digits and witness digits. To handle the MIL task, the model just needs to differentiate between the witness digit ("9") from other digits. Therefore, there should only be two clusters/classes.
Three clusters/classes are assigned to \textit{multiple digits} because the model needs to distinguish the two witness digits from the others.

For the case of \textit{outlier detection}, in order to detect the outlier(s) from a bag, the model needs to distinguish every digit.
For this reason, once capable of handling this MIL task, the models should also have the ability to cluster/classify the 10 digits.
It is clear that our model trained for this task has learned very good discriminative features for all 10-class digits, while the attention-based method fails, even when the mutual information loss is still applied on top of it.
The clustering accuracy is close to the known performance of ${\sim}98\%$ accuracy of the  supervised LeNet model~\cite{lecun-MNIST-2010}. This is strong evidence showing that our method is able to learn an instance-level representation in a weakly supervised manner.
In addition, Fig.~\ref{fig:tsneMNIST} shows the t-SNE visualizations for features extracted by our method in the testing bag of the four tasks. The figure clearly shows how discriminative the singleton features are. 

These results prove that our \textit{Hypothesis 1} is correct.

\begin{figure}[h]
\centering
\includegraphics[width=1\textwidth]{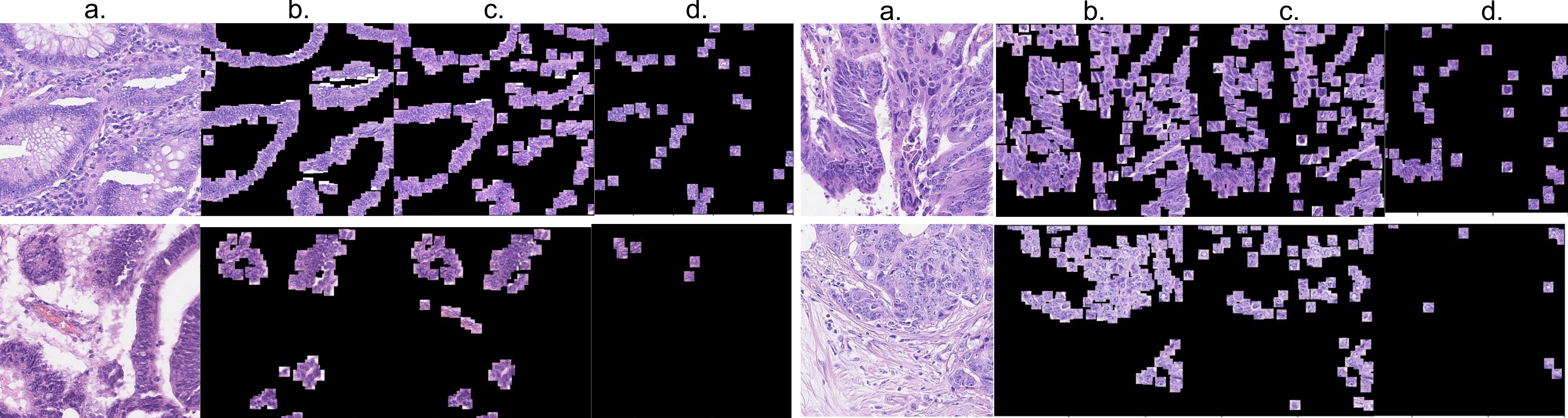}
\caption{a) The original H\&E image. b) The epithelial nuclei patches (Ground-Truth). c) The epithelial nuclei patches detected by our MIL model. d) The epithelial nuclei patches detected by attention-based MIL model}
\label{fig:colonCancerCompare}
\vspace{-2mm}
\end{figure}

\textbf{Instance-level prediction:}
% \subsubsection{Instance-level predictions:}
The colon cancer dataset contains 7,722 epithelial nuclei and 14,721 other nuclei. We select one of the models we trained earlier and treat the patches as singleton bags (i.e. bags only contain one patch). The singleton bags are sent to the model to make instance-level predictions:
epithelial or not. In the meantime we also use the same training-test split to train a fully supervised model. We report the instance-level accuracy in Table \ref{tab:instanceLabelcolonCancerExp}. In addition, Fig.~\ref{fig:colonCancerCompare} shows the patches that are classified as epithelial nuclei. 
% Our model has a better ability to get the instance-level label, which can be useful for pathologists.

%%--------------------------------------------------------------------------
% \begin{figure}[h]
% \centering
% \includegraphics[width=0.48\textwidth]{imgs/tsne_mnist_bag_v7.pdf}
% \caption{t-SNE visualizations of features extracted from \textit{Outlier detection} model.}
% \label{fig:tsneMNIST}
% \end{figure}

\textbf{Discussion}. This task meets the requirement of \textit{Hypothesis 2}: the bag representation $S$ contains the information whether the epithelial nuclei exist in a bag, which is close to what would be expected from instance-level feature $F$.
Our model achieves the best performance for bag-level prediction.
% In bag-level prediction, our model achieves the best performance among the baselines. 
It also has a good performance on the instance-level prediction. The mean accuracy is close to the supervised model and significantly better (${\sim}18$ pp) than that of the Attention-based model. It clearly shows that our MIL model can be used to predict the instance labels.
In addition, Fig.~\ref{fig:colonCancerCompare} shows that our model has a better ability to identify the nuclei of interest, which can be useful for pathologists.

% Please refer to the supplementary material for results of MNIST dataset experiment.
%%--------------------------------------------------------------------------
% \begin{table}
% \setlength{\tabcolsep}{4.7pt} % Default value: 6pt
% \centering
% \caption{Instance label accuracy for Colon cancer dataset}
% % \footnotesize
% \resizebox{0.5\columnwidth}{!}{
% \begin{tabular*}{10.5cm}
% {@{\extracolsep{\fill}} l c c c }
% \toprule 
% \multirow{2}{*}{Method} & TP  & TN & mean Acc \\ 
% &(test/train) & (test/train) & (test/train)\\
% \midrule[0.6pt]	
% Atten. & 32.42~/~21.25 &\textbf{98.45~/~99.22}& 65.43~/~63.60\\
% Ours & \textbf{73.47~/~70.73} & 92.39~/~92.28 &\textbf{82.93~/~81.51}\\
% Supervised & \textit{78.92~/~92.09} &\textit{91.14~/~98.22}& \textit{85.03~/~95.16}\\
% \bottomrule[1pt]
% \end{tabular*}
% }
% \label{tab:instanceLabelcolonCancerExp}
% % \vspace{-4mm}
% \end{table}
%%--------------------------------------------------------------------------

\vspace{-7mm}
%%%%%-------------------------------------------
\vspace{-12mm}

%%=============================================================================
%%  CONCLUSION
%%=============================================================================
% \vspace{-2mm}

\section{Conclusion}
\label{sec:conclusions}
% We presented a simple iterative approach to address MIL 
% problems.
% %
% Our method is capable of learning the underlying structure that characterizes 
% each of the bags by looking at its constituent elements one at a time.
% %
% Despite its simplicity the proposed method is able to effectively model 
% a variety of underlying MI assumptions and handle both classification 
% and regression task while requiring minimum modifications.
% %
% % A deeper analysis of the learned bag representation reveals that 
% % our method is able to highlight witness elements which are relevant to 
% % the underlying bag structure, thus, providing some explanation capabilities.
% {The proposed method can learn the instance-level distribution in a weakly supervised fashion.}
%
%%=====================================
% We investigate the potential of LSTMs at capturing information beyond order.
% % We conduct 
% Through an extensive analysis we have shown that LSTMs can indeed capture additional information of the underlying structure within the sets.
% % , such as the occurrence of instances and information accumulation. 
% Moreover, our results suggest that the performance at modeling more general bag structures is comparable and even better than that from MIL the methods specifically designed for this type problems.
% The proposed method can also model the instance-level distribution in a weakly supervised manner.
%%=====================================
\kaili{We investigate the potential of LSTMs at solving MIL problems.
Through an extensive analysis we have shown that LSTMs can indeed capture additional information of the underlying structure within the bags.
Our results suggest that the performance at modeling more general bag structures is comparable and even better than that from methods specifically designed for MIL problems.
The proposed method can also model the instance-level distribution in a weakly supervised manner.}

% \section{Related Work}
% \sect

\clearpage
\bibliographystyle{splncs}
\bibliography{egbib}

\begin{thebibliography}{10}

\bibitem{MILencyclopedia}
Sammut, C., Webb, G.I.:
\newblock Multi-instance learning.
\newblock Encyclopedia of Machine Learning (2011)

\bibitem{LSTM_97}
Hochreiter, S., Schmidhuber, J.:
\newblock Long short-term memory.
\newblock Neural Computation \textbf{9} (1997)  1735--1780

\bibitem{Sundermeyer_lstmneural}
Sundermeyer, M., Schlüter, R., Ney, H.:
\newblock Lstm neural networks for language modeling (2012)

\bibitem{SutskeverNIPS2014}
Sutskever, I., Vinyals, O., Le, Q.V.:
\newblock Sequence to sequence learning with neural networks.
\newblock In Ghahramani, Z., Welling, M., Cortes, C., Lawrence, N.D.,
  Weinberger, K.Q., eds.: Advances in Neural Information Processing Systems 27.
\newblock Curran Associates, Inc. (2014)  3104--3112

\bibitem{LeMikolovICML2014}
Le, Q.V., Mikolov, T.:
\newblock Distributed representations of sentences and documents.
\newblock In: ICML. JMLR Workshop and Conference Proceedings, JMLR.org (2014)

\bibitem{LSTM_action_recog_16}
Liu, J., Shahroudy, A., Xu, D., Wang, G.:
\newblock Spatio-temporal lstm with trust gates for 3d human action
  recognition.
\newblock In Leibe, B., Matas, J., Sebe, N., Welling, M., eds.: ECCV. (2016)

\bibitem{LSTM_CV_15}
SHI, X., Chen, Z., Wang, H., Yeung, D.Y., Wong, W.k., WOO, W.c.:
\newblock Convolutional lstm network: A machine learning approach for
  precipitation nowcasting.
\newblock In Cortes, C., Lawrence, N.D., Lee, D.D., Sugiyama, M., Garnett, R.,
  eds.: Advances in Neural Information Processing Systems 28.
\newblock Curran Associates, Inc. (2015)  802--810

\bibitem{Alahi_2016_CVPR}
Alahi, A., Goel, K., Ramanathan, V., Robicquet, A., Fei-Fei, L., Savarese, S.:
\newblock Social lstm: Human trajectory prediction in crowded spaces.
\newblock In: The IEEE Conference on Computer Vision and Pattern Recognition
  (CVPR). (2016)

\bibitem{zhu_aaai16}
Zhu, W., Lan, C., Xing, J., Zeng, W., Li, Y., Shen, L., Xie, X.:
\newblock Co-occurrence feature learning for skeleton based action recognition
  using regularized deep {LSTM} networks.
\newblock In: Thirtieth AAAI Conference on Artificial Intelligence. (2016)

\bibitem{Singh_2016_CVPR}
Singh, B., Marks, T.K., Jones, M., Tuzel, O., Shao, M.:
\newblock A multi-stream bi-directional recurrent neural network for
  fine-grained action detection.
\newblock In: The IEEE Conference on Computer Vision and Pattern Recognition
  (CVPR). (2016)

\bibitem{LSTM_speech_05}
Graves, A., Schmidhuber, J.:
\newblock Framewise phoneme classification with bidirectional lstm and other
  neural network architectures.
\newblock NEURAL NETWORKS (2005)  5--6

\bibitem{Gers99learningtoforget}
Gers, F.A., Schmidhuber, J., Cummins, F.:
\newblock Learning to forget: Continual prediction with lstm.
\newblock Neural Computation \textbf{12} (1999)  2451--2471

\bibitem{AmoresMILsurvey2013}
Amores, J.:
\newblock Multiple instance classification: Review, taxonomy and comparative
  study.
\newblock Artif. Intell. \textbf{201} (2013)  81--105

\bibitem{Carbonneau2018MultipleIL}
Carbonneau, M.A., Cheplygina, V., Granger, E., Gagnon, G.:
\newblock Multiple instance learning: A survey of problem characteristics and
  applications.
\newblock ArXiv \textbf{abs/1612.03365} (2018)

\bibitem{foulds_frank_2010}
Foulds, J., Frank, E.:
\newblock A review of multi-instance learning assumptions.
\newblock The Knowledge Engineering Review \textbf{25} (2010)  1–25

\bibitem{lecun-MNIST-2010}
LeCun, Y., Cortes, C.:
\newblock {MNIST} handwritten digit database.
\newblock (2010)

\bibitem{Lookbook}
Yoo, D., Kim, N., Park, S., Paek, A.S., Kweon, I.:
\newblock Pixel-level domain transfer.
\newblock ECCV (2016)

\bibitem{ITW:2018}
Ilse, M., Tomczak, J.M., Welling, M.:
\newblock Attention-based deep multiple instance learning.
\newblock arXiv preprint arXiv:1802.04712 (2018)

\bibitem{NTM14}
Graves, A., Wayne, G., Danihelka, I.:
\newblock Neural turing machines.
\newblock CoRR \textbf{abs/1410.5401} (2014)

\bibitem{graves2016hybrid}
Graves, A., Wayne, G., Reynolds, M., Harley, T., Danihelka, I.,
  Grabska-Barwińska, A., Colmenarejo, S.G., Grefenstette, E., Ramalho, T.,
  Agapiou, J., Badia, A.P., Hermann, K.M., Zwols, Y., Ostrovski, G., Cain, A.,
  King, H., Summerfield, C., Blunsom, P., Kavukcuoglu, K., Hassabis, D.:
\newblock Hybrid computing using a neural network with dynamic external memory.
\newblock Nature (2016)

\bibitem{fashionCompLSTM}
Han, X., Wu, Z., Jiang, Y.G., Davis, L.S.:
\newblock Learning fashion compatibility with bidirectional lstms.
\newblock In: ACM MM. MM ’17 (2017)

\bibitem{yazici2019orderless}
Yazici, V.O., Gonzalez-Garcia, A., Ramisa, A., Twardowski, B., van~de Weijer,
  J.:
\newblock Orderless recurrent models for multi-label classification (2019)

\bibitem{deepset17}
Zaheer, M., Kottur, S., Ravanbakhsh, S., Poczos, B., Salakhutdinov, R.R.,
  Smola, A.J.:
\newblock Deep sets.
\newblock In Guyon, I., Luxburg, U.V., Bengio, S., Wallach, H., Fergus, R.,
  Vishwanathan, S., Garnett, R., eds.: Advances in Neural Information
  Processing Systems 30.
\newblock Curran Associates, Inc. (2017)  3391--3401

\bibitem{RamonMINN}
Ramon, J., {De Raedt}, L.:
\newblock Multiple instance neural networks.
\newblock In: ICML Workshop on Attribute-value and Relational Learning. (2000)

\bibitem{WangDeepMIL}
:
\newblock Revisiting multiple instance neural networks.
\newblock Pattern Recognition \textbf{74} (2018)  15–24

\bibitem{LiuCVPR17attention}
Liu, Y., Yan, J., Ouyang, W.:
\newblock Quality aware network for set to set recognition.
\newblock (2017)

\bibitem{yan18DynamicPooling}
Yan, Y., Wang, X., Guo, X., Fang, J., Liu, W., Huang, J.:
\newblock Deep multi-instance learning with dynamic pooling.
\newblock In: Asian Conference on Machine Learning. (2018)

\bibitem{tiboMMIN}
Tibo, A., Jaeger, M., Frasconi, P.:
\newblock Learning and interpreting multi-multi-instance learning networks.
\newblock arXiv:1810.11514 (2018)

\bibitem{MingMIGNN}
Tu, M., Huang, J., He, X., Zhou, B.:
\newblock Multiple instance learning with graph neural networks.
\newblock In: ICML Workshop on Learning and Reasoning with Graph-Structured
  Representations. (2019)

\bibitem{VGG}
Simonyan, K., Zisserman, A.:
\newblock Very deep convolutional networks for large-scale image recognition.
\newblock arXiv:1409.1556 (2014)

\bibitem{He2015DeepRL}
He, K., Zhang, X., Ren, S., Sun, J.:
\newblock Deep residual learning for image recognition.
\newblock In: CVPR. (2015)

\bibitem{MikolovWord2Vec}
Mikolov, T., Sutskever, I., Chen, K., Corrado, G.S., Dean, J.:
\newblock Distributed representations of words and phrases and their
  compositionality.
\newblock In Burges, C.J.C., Bottou, L., Welling, M., Ghahramani, Z.,
  Weinberger, K.Q., eds.: Advances in Neural Information Processing Systems 26.
\newblock Curran Associates, Inc. (2013)  3111--3119

\bibitem{devlinBERT}
Devlin, J., Chang, M., Lee, K., Toutanova, K.:
\newblock {BERT:} pre-training of deep bidirectional transformers for language
  understanding.
\newblock In: arXiv:1810.04805. (2018)

\bibitem{FernandoAl:TPAMI16}
Fernando, B., Gavves, E., Oramas~M., J., Ghodrati, A., Tuytelaars, T.:
\newblock Rank pooling for action recognition.
\newblock In: TPAMI. (2016)

\bibitem{bilenDynamicImages}
{Bilen}, H., {Fernando}, B., {Gavves}, E., {Vedaldi}, A.:
\newblock Action recognition with dynamic image networks.
\newblock TPAMI \textbf{40} (2018)  2799--2813

\bibitem{infomax2018}
Hjelm, R.D., Fedorov, A., Lavoie-Marchildon, S., Grewal, K., Bachman, P.,
  Trischler, A., Bengio, Y.:
\newblock Learning deep representations by mutual information estimation and
  maximization.
\newblock In: International Conference on Learning Representations. (2019)

\bibitem{YangDualAEN2019}
Yang, X., Deng, C., Zheng, F., Yan, J., Liu, W.:
\newblock Deep spectral clustering using dual autoencoder network.
\newblock CVPR (2019)

\bibitem{lenet}
{Lecun}, Y., {Bottou}, L., {Bengio}, Y., {Haffner}, P.:
\newblock Gradient-based learning applied to document recognition.
\newblock Proceedings of the IEEE (1998)

\bibitem{LSTM_counting_19}
Suzgun, M., Gehrmann, S., Belinkov, Y., Shieber, S.M.:
\newblock {LSTM} networks can perform dynamic counting.
\newblock ACL 2019 Workshop on Deep Learning and Formal Languages (2019)

\bibitem{DeepFashion}
Liu, Z., Luo, P., Qiu, S., Wang, X., Tang, X.:
\newblock Deepfashion: Powering robust clothes recognition and retrieval with
  rich annotations.
\newblock In: Proceedings of IEEE Conference on Computer Vision and Pattern
  Recognition (CVPR). (2016)

\bibitem{multipleQueries}
Arandjelovi\'c, R., Zisserman, A.:
\newblock Multiple queries for large scale specific object retrieval.
\newblock In: BMVC. (2012)

\bibitem{Ricci-Vitiani2007}
Ricci-Vitiani, L., Lombardi, D.G., Pilozzi, E., Biffoni, M., Todaro, M.,
  Peschle, C., De~Maria, R.:
\newblock Identification and expansion of human colon-cancer-initiating cells.
\newblock Nature \textbf{445} (2007)  111--115

\bibitem{colon_cancer_2016}
{Sirinukunwattana}, K., {Raza}, S.E.A., {Tsang}, Y., {Snead}, D.R.J., {Cree},
  I.A., {Rajpoot}, N.M.:
\newblock Locality sensitive deep learning for detection and classification of
  nuclei in routine colon cancer histology images.
\newblock IEEE Transactions on Medical Imaging (2016)

\bibitem{LiangICCV17}
Liang, X., Lee, L., Dai, W., Xing, E.P.:
\newblock Dual motion gan for future-flow embedded video prediction.
\newblock In: International Conference on Computer Vision. Volume
  abs/1708.00284. (2017)  1762--1770

\bibitem{tsne}
van~der Maaten, L., Hinton, G.:
\newblock Visualizing high-dimensional data using t-sne.
\newblock Journal of Machine Learning Research (2008)

\end{thebibliography}
\end{document}